\pdfoutput=1

\documentclass[11pt]{article}

\usepackage{acl}

\usepackage{times}
\usepackage{latexsym}
\usepackage{amsmath}
\usepackage{multirow}
\usepackage[T1]{fontenc}
\usepackage{xcolor,pifont}

\usepackage[utf8]{inputenc}

\usepackage{microtype}

\usepackage{inconsolata}

\usepackage{graphicx}

\title{WinClick: GUI Grounding with Multimodal Large Language Models}

\author{
    \textbf{Zheng Hui\textsuperscript{$\clubsuit$ $\heartsuit$}},
    \textbf{Yinheng Li\textsuperscript{$\clubsuit$}},
    \textbf{Dan zhao\textsuperscript{$\clubsuit$}}, 
    \textbf{Tianyi Chen\textsuperscript{$\clubsuit$}},
    \\
    \textbf{Colby Banbury\textsuperscript{$\clubsuit$}},
    \textbf{Kazuhito Koishida\textsuperscript{$\clubsuit$}}
    \\
    \textsuperscript{$\clubsuit$}Microsoft,
    \textsuperscript{$\heartsuit$}Columbia University
    \\
    \{yinhengli, tianyi.chen, colbybanbury \}@microsoft.com,
    \\
    \{zh2483\}@columbia.edu, \{dz1158\}@nyu.edu
}

\begin{document}

\maketitle

\begin{abstract}

Graphical User Interface (GUI) tasks are vital for automating workflows such as software testing, user interface navigation. For users, the GUI is the most intuitive platform for interacting with a computer. Previous work identified a key challenge in developing visual GUI agents: GUI grounding—the ability to accurately locate screen elements based on instructions. However, most existing GUI agents rely on structured data formats like DOM or HTML files in training or inferencing, which are inaccessible across all applications, particular in a general desktop environments such as Windows OS. To address this, we introduce \textbf{WinClick}, a novel visual GUI agent developed in Windows platform. \textbf{WinClick} leverages screenshots to detect actionable regions. To overcome the challenge of GUI grounding, we enhance \textbf{WinClick} with GUI grounding pre-training and propose an LLM-based method for aligning GUI grounding data. Additionally, we introduce \textbf{WinSpot}, the first comprehensive benchmark for GUI grounding on Windows. Our experiments demonstrate that \textbf{WinClick}, combined with GUI grounding pre-training, significantly outperforms existing baselines, offering a scalable solution for GUI automation in desktop environments. \footnote{ https://github.com/zackhuiiiii/WinSpot.}
\end{abstract}
\section{Introduction}

With the emergence of large language models \cite{brown2020languagemodelsfewshotlearners,openai2024gpt4technicalreport}, various Multimodal Large Language Models (MLLMs) have demonstrated powerful capabilities \cite{liu2023visualinstructiontuning, bai2023qwenvlversatilevisionlanguagemodel,abdin2024phi3technicalreporthighly, openai2024gpt4technicalreport}.  Especially these models have demonstrated powerful understanding and reasoning capabilities \cite{NEURIPS2023_2b9efb08,liu-etal-2024-mmc,zhou-etal-2023-rome}, helping to automate interactions with Graphical User Interfaces (GUIs) across industries \cite{yang2023setofmarkpromptingunleashesextraordinary,li-etal-2020-mapping}. GUI agents have become essential tools in managing increasingly complex software systems, widely used for tasks such as automated software testing, user interface navigation, and application management \cite{wang2024survey}. By reducing manual labor, GUI agents improve efficiency and scalability, particularly in handling repetitive or intricate tasks\cite{zhou2023webarena, NEURIPS2023_5950bf29, pmlr-v235-zheng24e}.
Those GUI agents rely on structured data formats, which makes them well-suited for web environments where such data is readily available \cite{kim2024language, gur2024realworldwebagentplanninglong}.  

\begin{figure}[t]
  \begin{center}
  \includegraphics[width=1\columnwidth]{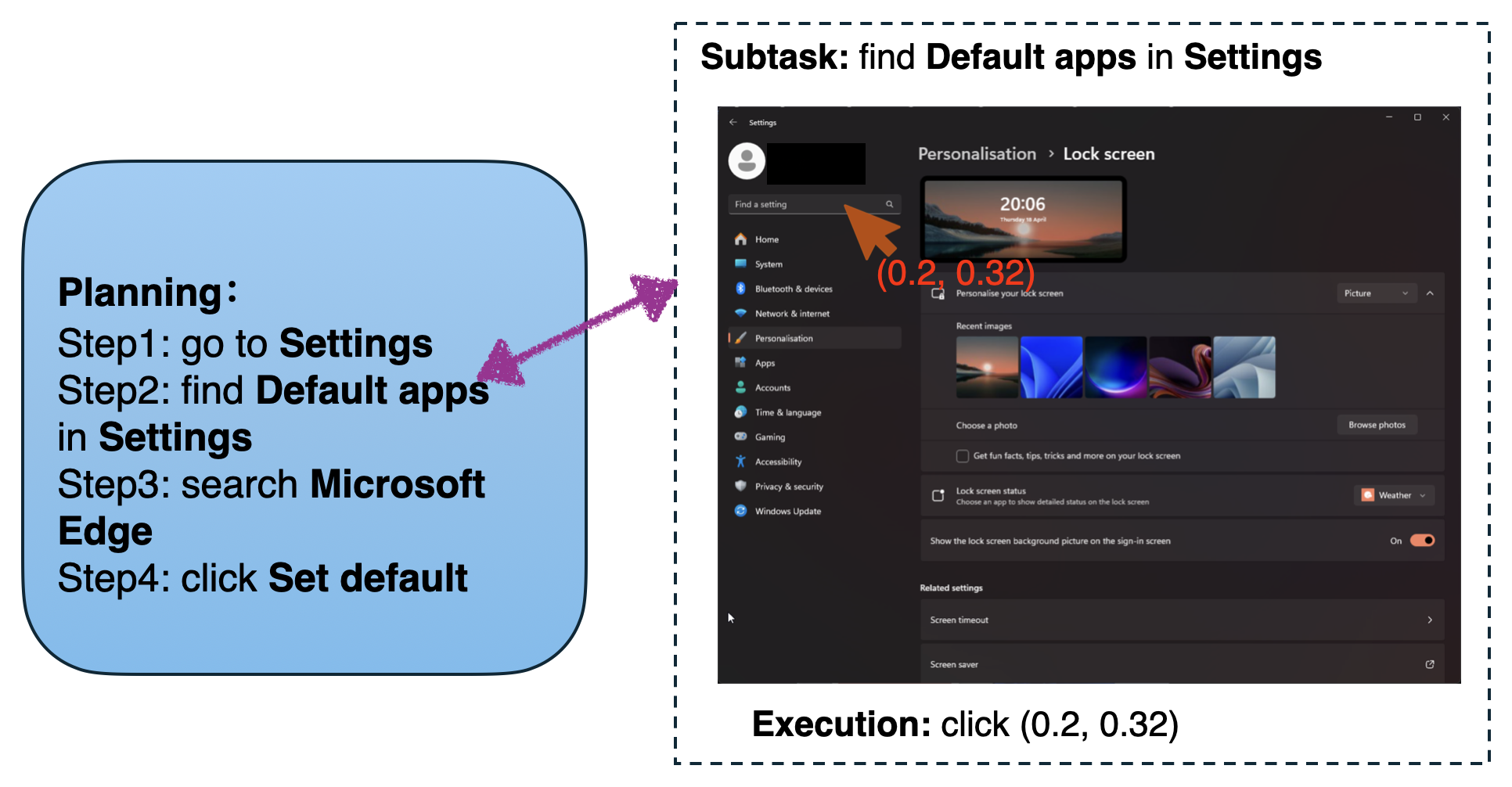}
  \end{center}
  \captionsetup{justification=centering}
  \caption{GUI Agent subtask, GUI grounding }
  \label{fig:f1}
\end{figure}

However, desktop environments pose different challenges as structured data is often inaccessible, leaving only visual information, such as screenshots, as the primary source for interaction. The development of GUI agents for these environments poses a critical question: how can agents operate effectively when they cannot access structured data? To address this, recent advances \cite{bavishi2023fuyu, yang2023setofmarkpromptingunleashesextraordinary, cheng-etal-2024-seeclick} have attempted to bypass the need for structured text by leveraging MLLMs to process screenshots directly. While this approach has seen success in mobile and web-based platforms \cite{wang2024mobile,cheng-etal-2024-seeclick}, it has largely overlooked desktop environments. 

One of the challenges in developing visual-based GUI agents is GUI grounding—the ability to accurately identify and interact with elements on the screen based solely on visual inputs as shown in Fig \ref{fig:f1}. Prior work \cite{wang2024mobile, cheng-etal-2024-seeclick} in this space has made significant progress with web and mobile interfaces, but it does not focus on desktop environments, especially Windows. Furthermore, many of these methods rely on LLMs (>10B) that are inefficient for real-time desktop applications, and there is no comprehensive benchmark for evaluating GUI grounding performance specifically on Windows platforms.

To bridge this gap, we propose \textbf{WinClick}, a novel GUI agent that developed in Windows environments and potentially can be used across different OS platforms. WinClick relies solely on visual data for automating tasks. WinClick leverages screenshots to detect actionable regions, performing GUI grounding without the need for structured data formats during training or inference. Our model is based on the Phi3-vision model, which is considered a "small" MLLM and suitable for on-device use. By focusing on desktop environments, WinClick addresses a major limitation in previous work, offering a solution tailored to the Windows operating system.

To enhance the performance of WinClick, we introduce a novel image and description alignment method for automating the curation of GUI grounding data using MLLMs, enabling more efficient training. Additionally, to fill the void in benchmarking, we present WinSpot, the first benchmark designed specifically for GUI grounding tasks in Windows. WinSpot includes over 1,000 images and 5,000 instruction-click pairs, providing a Windows-specific benchmark for GUI agents.

Our experimental results demonstrate that WinClick, equipped with GUI grounding pre-training, outperforms or is comparable with baselines on benchmarks, validating the efficacy of our approach. The main contributions of this paper are as follows:

\begin{itemize}
    \item We introduce \textbf{WinClick}, a visual GUI agent specifically designed for Windows environments that enhance the ability to locate and interact with screen elements using visual data alone, addressing a fundamental challenge in GUI automation.
    
    \item We develop an automated method using MLLMs to align GUI grounding data, ensuring accurate mapping between visual elements and instructions, while also reducing the effort required for human labeling.
    
    \item We introduce \textbf{WinSpot}, the first comprehensive benchmark for GUI grounding in Windows environments.
    
    \item We demonstrate that \textbf{WinClick}, under GUI alignment training, outperforms baselines on the \textbf{WinSpot} benchmark, confirming the importance of GUI grounding in enhancing agent performance on desktop platforms.    
\end{itemize}

\begin{figure*}[htbp]
  \begin{center}
  \includegraphics[width=1.89\columnwidth]{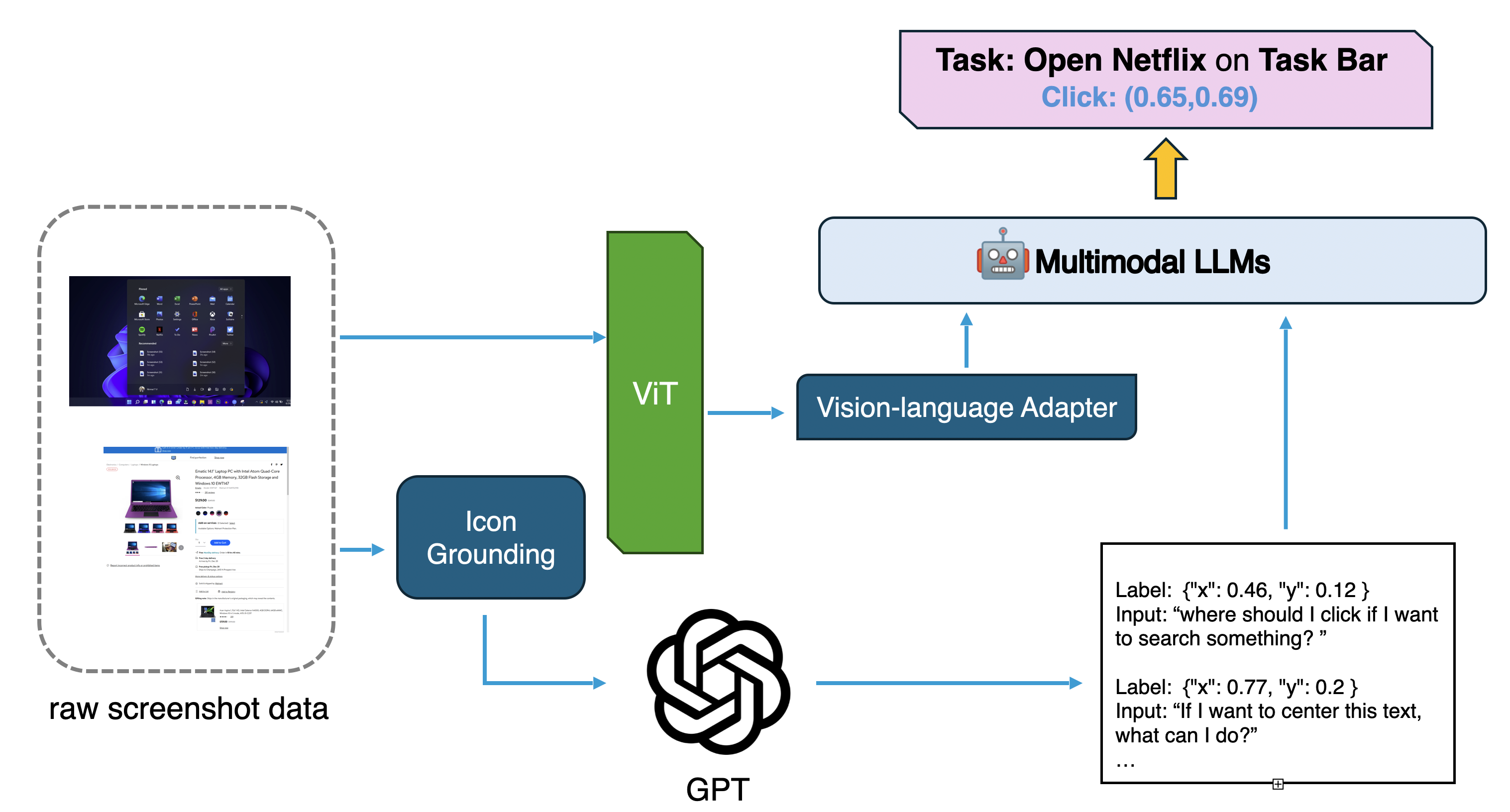}
  \end{center}
  \captionsetup{justification=centering}
  \caption{WinClick's full workflow}
  \label{fig:f2}
\end{figure*}

\section{Related Work}

\subsection{UI Screen Understanding Dataset}

A variety of datasets \cite{moran2018machine, he2021actionbert, wu2023webui} have been developed to support UI modeling, primarily in the mobile domain. For instance, the AMP dataset \cite{zhang2021screen}, containing 77k screens from 4,068 iOS apps. Another significant resource is Rico \cite{deka2017rico}, the largest publicly available dataset for UI understanding. Rico includes 72k app screens from 9.7k Android apps, collected through a mix of automated and human crawling methods. 
In the broader web and OS domain, datasets such as Mind2Web \cite{deng2024mind2web} , Visual-WebArena \cite{koh2024visualwebarena}, and Windows Arena \cite{bonatti2024windows} offer simulated environments for various tasks. However, these datasets do not explicitly address GUI challenges but more focus on downstream agent work. Despite the breadth of available data, a critical gap remains: the lack of comprehensive desktop-specific UI datasets, particularly for the Windows operating system.
Existing datasets overwhelmingly focus on mobile and web platforms, leaving desktop environments underexplored. The closest dataset related to desktop UI understanding is SeeClick \cite{cheng-etal-2024-seeclick}, though it predominantly targets cross-platform settings and lacks a specific focus on Windows. Our work fills this gap by introducing a dataset tailored to desktop environments, particularly for Windows OS, which marks a novel contribution to the field.

\subsection{Autonomous GUI Navigation and MLLMs}

With the rise of LLMs like the GPT series \cite{brown2020languagemodelsfewshotlearners}, there has been a shift toward using these models for more complex GUI understanding or automation. However, LLMs are limited in handling visual input, which is essential for many GUI tasks.
Recent efforts have explored vision-based GUI agents using multimodal LLMs , such as GPT-4V \cite{yang2023setofmarkpromptingunleashesextraordinary, bonatti2024windows}, which combine visual and textual data to navigate UIs.
Current approaches to GUI agents either predict actions directly, as seen in models like \citet{shaw2023pixels}, or use multimodal models like GPT-4V for UI tasks, as demonstrated by \citet{yang2023setofmarkpromptingunleashesextraordinary, bonatti2024windows}. While these methods show promise, many still rely on structured data like HTML or DOM trees in either training or inferencing, limiting their applicability in desktop environments where such data is often unavailable.
In contrast, our work offers a more general solution from training to inferencing without using HTML or Dom files.

\section{Method}

\begin{figure*}[htbp]
  \begin{center}
  \includegraphics[width=1.9\columnwidth]{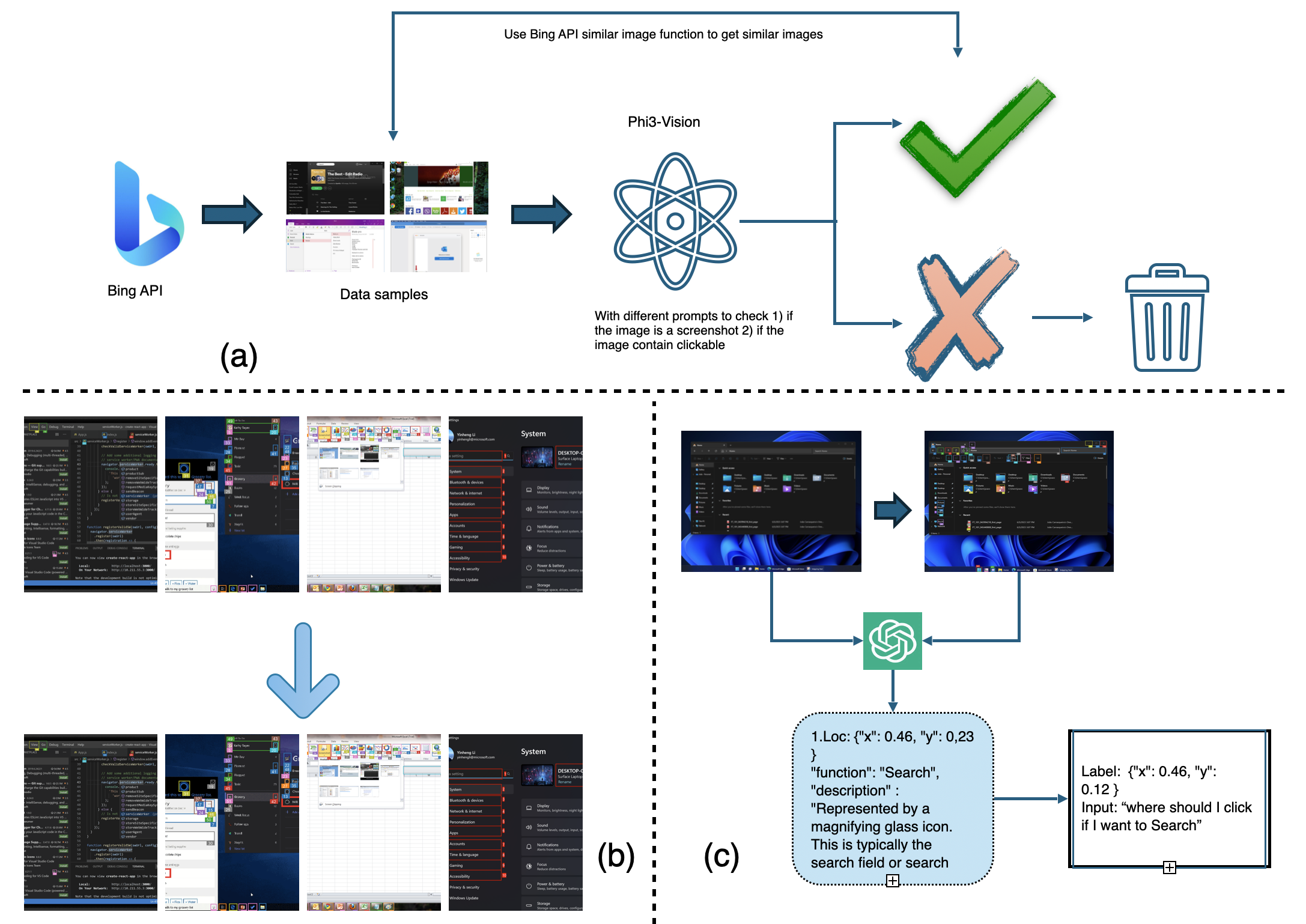}
  \end{center}
  \captionsetup{justification=centering}
  \caption{Overview of the Data Alignment Process: (a) illustrates the data filtering strategy using the Phi3 vision model, (b) shows the input and output of icon grounding with the in-house ViT-BERT model, and (c) demonstrates the use of LLMs for GUI and description alignment.}

  \label{fig:f4}
\end{figure*}

While recent MLLMs have demonstrated grounding capabilities on natural images, GUI screenshots present a different set of challenges. These screenshots, especially in desktop environments like Windows, often feature densely packed text, icons, and widgets, complicating the grounding process and limiting the performance of existing VLMs in GUI contexts.
To address these limitations, we introduce WinClick, a visual GUI agent specifically designed for Windows environments. This section details the architecture and methodology behind WinClick, including how we formalize the GUI grounding task, pre-training data alignments, Winspot dataset and how to train/eval the model.

\subsection{GUI Grounding for MLLMs}

In GUI automation, the challenge lies in grounding—accurately mapping elements based on visual input and instructions. Let \( S \) represent an interface screenshot, and \( \mathcal{E} = \{e_i = (x_i, y_i)\ | i = 1, \dots, N \} \) denote a set of GUI elements, where \( x_i \) is the textual description and \( y_i \) is the corresponding location (either as a bounding box or a point). 
The task can be framed as two-fold, 1) Forward task: Estimating \( P(y | S, x) \), the probability that the location \( y \) corresponds to the element \( e \), given the screenshot \( S \) and description \( x \). 2) Reverse task: Estimating \( P(x | S, y) \), the probability that description \( x \) corresponds to a location \( y \), given the screenshot \( S \).

Previous works \cite{shaw2023pixels} tackled this by dividing the image into 1,000 bins and creating a 1,000-token vocabulary to represent x and y coordinates. Instead, WinClick adopts a more intuitive approach from \cite{cheng-etal-2024-seeclick} by treating coordinates as natural language, avoiding additional tokenization. For instance, when given an instruction like "Click the 'Settings' icon," we generate a prompt such as "I want to kill this program, where should I click to \textit{<instruction>}?" and predict the coordinates \( (0.52, 0.35) \).
The training objective involves minimizing the loss for both tasks. For the forward task, the cross-entropy loss between the predicted coordinates \( \hat{y} \) and the ground truth \( y \) is:

\small
\[
\mathcal{L}_{\text{grounding-forward}} = - \sum_{i=1}^{N} \left( y_i \log \hat{y}_i + (1 - y_i) \log (1 - \hat{y}_i) \right)
\]
\normalsize

For the reverse task, the cross-entropy loss between the predicted description \( \hat{x} \) and the ground truth \( x \) is:

\small
\[
\mathcal{L}_{\text{grounding-reverse}} = - \sum_{i=1}^{N} \left( x_i \log \hat{x}_i + (1 - x_i) \log (1 - \hat{x}_i) \right)
\]
\normalsize

\subsection{Data Construction}
\label{section:data}

\begin{figure}[ht]
  \begin{center}
  \includegraphics[width=1\columnwidth]{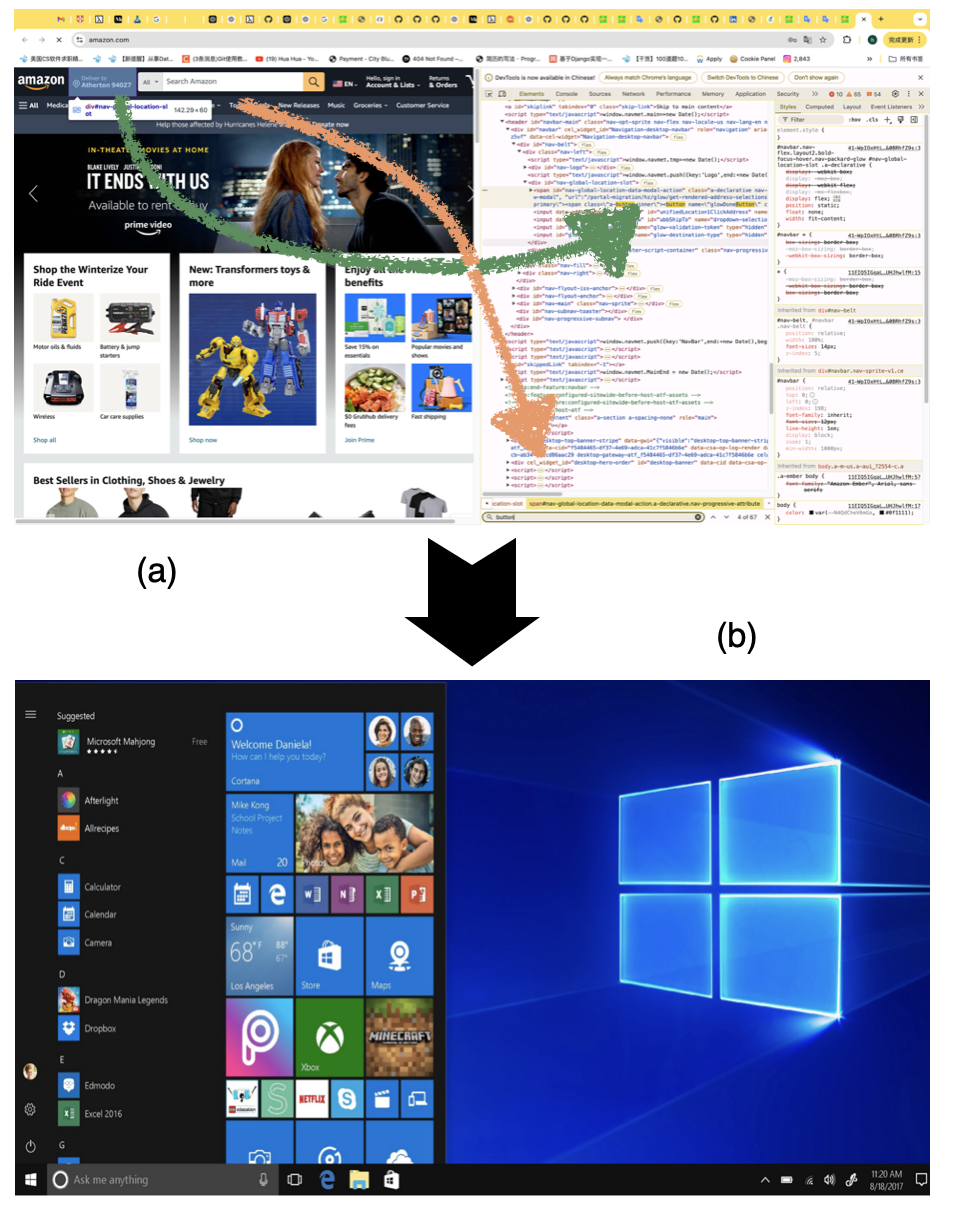}
  \end{center}
  \captionsetup{justification=centering}
  \caption{(a) Traditional methods rely on HTML or DOM files to locate icons during data construction. (b) Our proposed data alignment framework requires only raw screenshot images.}
  \label{fig:f3}
\end{figure}

Unlike previous work \cite{cheng-etal-2024-seeclick}, which focuses on cross-domain tasks and using structured data for constructing training datasets, our work is centered on the Windows system (as shown in Figure \ref{fig:f3}. To address the unique challenges of this environment, we designed a new Instruction-Interactable Region Alignment method as shown in Figure \ref{fig:f4}, which leverages MLLMs to generate high-quality training data without relying on HTML elements or web-based data.

We begin by searching and filtering images from the web using the Bing API, followed by a quality check using the smaller phi3-vision model. The gold is to collect a list of diverse and representative screenshot in Windows. So, we query screenshots for 700+ most popular apps listed in Microsoft stores \footnote{https://www.microsoft.com/en-us/store/most-popular/apps/pc}. This model primarily checks the resolution and ensures the image is a valid screenshot. If the image meets the quality criteria, we randomly add it to our data bank. We further enhance the dataset by using the Bing API’s similar image functionality to add visually related images to our candidate pool. For images that do not meet the resolution or quality threshold, we discard them.

Once filtered, we apply a proprietary Bert model with ViT encoder to perform icon grounding on the selected images. The ViT-Bert model generates bounding boxes around interactable icons in the images, which we use to create structured data. We then use GPT-4o for aligning the filtered images with corresponding icon descriptions. This alignment process serves multiple purposes: 1) By using expensive models like GPT-4o only in the data alignment stage, we reduce computational costs while maintaining accuracy during the reasoning and inference processes. 2) Previous work \cite{zheng2024gpt4visiongeneralistwebagent} has shown that providing GPT-4V with screenshots containing only bounding boxes and IDs can mislead the model. GPT-4V’s limited ability to simultaneously identify semantic information and predict actions creates a challenge. To mitigate this, we introduce semantic information directly into the images during data construction, applying a form of knowledge distillation to a smaller model. 3) By enriching the dataset with diverse semantic information, we ensure that the subsequent click agents can handle distributed tasks more effectively, improving overall performance.
In addition to the data collected via the Bing API, we incorporated 500 images from CoVA dataset\cite{kumar-etal-2022-cova}, 500 images from WebSight dataset\cite{laurençon2024unlocking} to further enhance our training set. Result in a dataset around 60K to train our model. For more examples about the data construction, please refer to Appendix \ref{moreexample}.

\begin{figure}[h]
  \begin{center}
  \includegraphics[width=1\columnwidth]{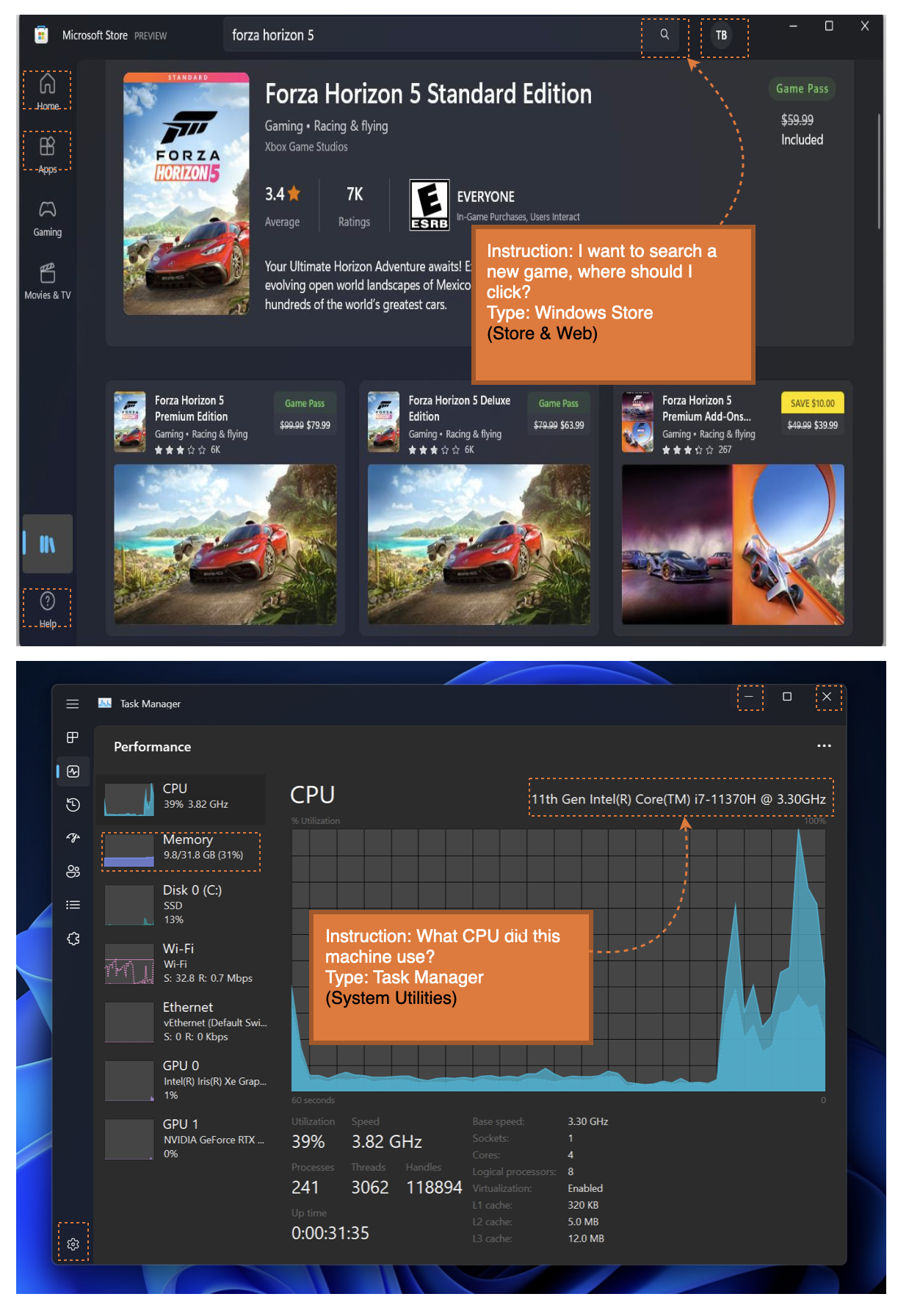}
  \end{center}
  \captionsetup{justification=centering}
  \caption{WinSpot examples}
  \label{fig:winspot}
\end{figure}

\subsection{WinSpot Benchmark}

To evaluate the effectiveness of GUI grounding and interaction in Windows environments, we introduce \textit{WinSpot}, the first comprehensive benchmark specifically designed for Windows desktop applications. While previous work \cite{cheng-etal-2024-seeclick} focused on cross-domain GUIs, \textit{WinSpot} is tailored for tasks and interfaces unique to the Windows ecosystem. It covers a broad range of common Windows applications, making it ideal for testing the robustness and generalization of visual GUI agents like WinClick.

The \textit{WinSpot} dataset consists of over 1,000 annotated\footnote{More annotation detail in Appendix \ref{human}} screenshots from 14 core Windows applications, each representing unique interaction types and layout structures. These applications are widely used in both personal and professional contexts, ensuring the dataset captures a broad spectrum of real-world scenarios. The applications and their respective contributions to the dataset showed in Figure \ref{fig:f3}.
Each screenshot in \textit{WinSpot} contains multiple interactable regions, such as buttons, menus, and icons, each annotated with its corresponding function. These annotations include bounding boxes around the interactable elements and their associated semantic descriptions, which are aligned with natural language instructions for both grounding and task prediction tasks. This variety ensures that \textit{WinSpot} provides a challenging and comprehensive evaluation framework for GUI agents, enabling robust testing of both interaction precision and generalization across different applications.
\textit{WinSpot} presents a diverse array of tasks, including file navigation, system settings adjustment, and text input, as well as more specialized tasks such as process management in Task Manager and command execution in Command Prompt. These tasks encompass a wide range of complexity, from simple button clicks to more involved interactions that require an understanding of application-specific layouts.
In addition to supporting GUI grounding tasks \( P(y | S, x) \), \textit{WinSpot} also be used in reverse tasks \( P(x | S, y) \), where the model must predict the description of a given GUI element based on its location in the screenshot. This two-way task formulation enhances the evaluation by testing both the agent’s understanding of visual cues and its ability to map interactions to the correct interface components.

\begin{figure}[htbp]
  \begin{center}
  \includegraphics[width=1\columnwidth]{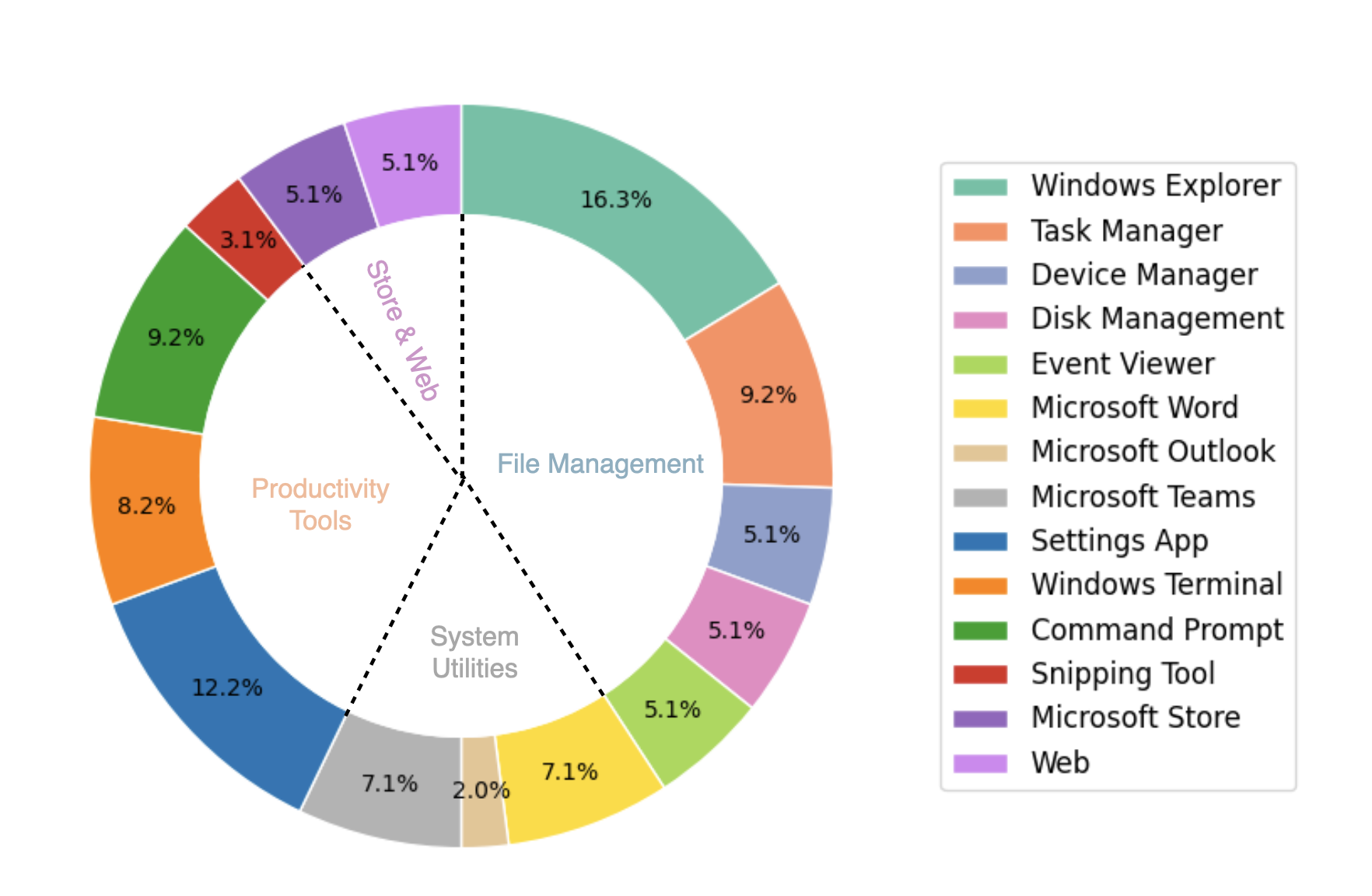}
  \end{center}
  \captionsetup{justification=centering}
  \caption{WinSpot Category}
  \label{fig:dountchart}
\end{figure}

\noindent With its focus on real-world Windows applications, \textit{WinSpot} fills a critical gap in the existing landscape of GUI agent evaluation, providing a benchmark that is both realistic and challenging. It ensures that agents trained on this dataset are capable of handling diverse and practical tasks in actual desktop environments.

\newcommand*\colourcheck[1]{%
  \expandafter\newcommand\csname #1check\endcsname{\textcolor{#1}{\ding{52}}}%
}
\colourcheck{green}
\newcommand*\colourxmark[1]{%
  \expandafter\newcommand\csname #1xmark\endcsname{\textcolor{#1}{\ding{56}}}%
}
\colourxmark{red}

\begin{table*}[h]
\centering
\renewcommand{\arraystretch}{1.2}
\scalebox{0.85}{\begin{tabular}{lcccccccc}
\hline

\multirow{2}{*}{\textbf{Method}}        & \multirow{2}{*}{\textbf{Size}} & \multirow{2}{*}{\textbf{GUI Specific}} & \multirow{2}{*}{\textbf{HTML or Dom}} & \multicolumn{2}{c}{\textbf{ScreenSpot}} & \multirow{2}{*}{\textbf{WinSpot}} & \multirow{2}{*}{\textbf{Avg}} \\ \cline{5-6}
                       &               &                    &     & Desktop-Text & Desktop-Icon  & \\ \hline
\textbf{MiniGPT-v2}    & 7B            & No \redxmark     & No \redxmark  & 6.2\%     & 2.9\%  & 1.7\% & 3.6\% \\ 
\textbf{GPT-4V}        & Unknown       & No \redxmark     & No \redxmark  & 20.2\%    & 11.8\% & 18.3\% & 16.8\% \\ 
\textbf{GPT-4o}        & Unknown       & No \redxmark     & No \redxmark  & 35.5\%    & 12.9\% & 16.5\% & 21.6\% \\ 
\textbf{Phi-3 Vision}  & \textbf{4.2B}          & No \redxmark     & No \redxmark  & 10.0\%    & 4.3\%  & 5.9\% & 6.7\% \\ \hline
\textbf{Fuyu}          & 8B            & Yes \greencheck       & No \redxmark  & 33.0\%    & 3.6\%  & 9.4\% & 15.3\% \\ 
\textbf{CogAgent}      & 18B           & Yes \greencheck        & No \redxmark  & \textbf{74.2\%}    & 20.0\% & 13.3\% & 35.8\% \\ 
\textbf{SeeClick}      & 9.6B          & Yes \greencheck        & No \redxmark  & \underline{72.2\%}    & \underline{30.0\%} & 15.7\% & 39.3\% \\ 
\textbf{Ours (LoRA)}   & \textbf{4.2B}          & Yes \greencheck        & Yes \greencheck     & 60.4\%    & 29.1\% & \underline{47.3\%} & \underline{45.6\%} \\
\textbf{Ours (Full)}   & \textbf{4.2B}          & Yes \greencheck        & Yes \greencheck     & 66.5\%    & \textbf{45.6\%} & \textbf{56.2\%} & \textbf{56.1\%} \\ \hline
\end{tabular}}
\caption{Evaluation of various methods on ScreenSpot and WinSpot. 'HTML or DOM' indicates whether the method is HTML or DOM free for training data construction. 'GUI Specific' refers to whether the model is fine-tuned or trained specifically for GUI tasks. Some results are taken from \citet{cheng-etal-2024-seeclick} due to computational limitations. Bold indicates the best-performing result, while underscore indicates the second-best.}

\label{table:results_whole}
\end{table*}

\section{Experiments and Results}

In this section, we first describe the training details of WinClick. We then evaluate the GUI grounding capabilities of WinClick compared to other baseline models using WinSpot benchmark. Lastly, we evaluate our WinClick on other dataset.

\subsection{Training Details}
We develop WinClick through continual pre-training on the Phi3-vision model, a relatively smaller yet highly suitable model for PC environments. We train Phi3-vision on our custom dataset (as outlined in Section \ref{section:data}) for 3 epochs to create the base GUI model, WinClick. While we primarily employ full fine-tuning of both the visual encoder and language model during training, we also experiment with parameter-efficient methods such as LoRA and QLoRA to assess potential improvements in efficiency. For more detail, please refer to Appendix \ref{train}.

\subsection{Evaluation Datasets}

As noted in previous studies \cite{cheng-etal-2024-seeclick}, GUI grounding, a critical component for visual GUI agents, has not received sufficient attention in current MLLM evaluations\cite{liu2023mmbench, OSWorld}. Prior work in this area is limited, with ScreenSpot \cite{cheng-etal-2024-seeclick} being the only existing GUI-specific benchmark.  Thus, to thoroughly assess the performance of our models and baseline models, we employ two datasets: ScreenSpot and our newly introduced WinSpot. 

\paragraph{ScreenSpot-Desktop:} As the only existing GUI-specific benchmark, ScreenSpot-Desktop provides a baseline for evaluating GUI grounding in desktop environments. The Desktop part contains 334 images with description and location pair.

\paragraph{WinSpot:} In addition, we introduce WinSpot, a comprehensive benchmark specifically tailored for Windows environments. Inlcude 1000+ images and 5000+ description and location pairs.

\subsection{Baseline models}

We conducted evaluations on a variety of models, including both generalist MLLMs and those specifically designed for GUI tasks. For the general models, we tested GPT-4o \cite{openai2024gpt4technicalreport}, GPT-4V \cite{openai2024gpt4technicalreport}, Phi3-Vision \cite{abdin2024phi3technicalreporthighly}, and MinGPT-v2 \cite{chen2023minigptv2largelanguagemodel}. In addition, we evaluated more recent models tailored for GUI agent tasks, such as Fuyu \cite{bavishi2023fuyu}, CogAgent \cite{wang2024cogvlmvisualexpertpretrained}, and SeeClick \cite{cheng-etal-2024-seeclick}, which have been specifically designed to handle the complexities of GUI interaction.
In line with prior studies \cite{cheng-etal-2024-seeclick}, we define our evaluation criteria based on the requirements of GUI agents. For this, we use click accuracy as the key metric, measuring the percentage of instances where the model's predicted click location correctly falls within the boundaries of the ground truth element’s bounding box \cite{li2022grounded, yang2023dawn}.

\subsection{Results}

Table~\ref{table:results_whole} presents the evaluation of various models on the ScreenSpot and WinSpot benchmarks. General-purpose models like MiniGPT-v2 (7 billion parameters), GPT-4V, GPT-4o, and Phi-3 Vision (4.2 billion parameters) exhibit varying performance. MiniGPT-v2, a smaller general model, underperforms significantly, with an average of 3.6\%, showing its limitations in complex GUI tasks. Larger models like GPT-4V and GPT-4o show stronger performance, with GPT-4o achieving an average of 21.6\%, although they still fall short when it comes to the highly specialized grounding tasks required by the benchmarks. Phi-3 Vision, a smaller but more efficient model with only 4.2 billion parameters, achieves a modest 6.7\% average, highlighting the need for specialized fine-tuning in handling GUI-specific tasks.

When comparing WinClick to its base model, Phi-3 Vision, the difference in performance is striking. Phi-3 Vision scores only 10.0\% on Desktop-Text, 4.3\% on Desktop-Icon, and 5.9\% on WinSpot, with an average of 6.7\%. In contrast, WinClick, which is built on Phi-3 but fine-tuned specifically for GUI interaction tasks, shows a significant improvement. Even in the parameter-efficient LoRA version of WinClick, the model achieves an average of 45.6\%, with a much stronger performance across all benchmarks. The full fine-tuned version of WinClick performs even better, with an average score of 56.1\%, outperforming Phi-3 Vision across all metrics. The improvements in Desktop-Text (from 10.0\% to 66.5\%) and WinSpot (from 5.9\% to 56.2\%) are particularly noteworthy, indicating that WinClick's specialized training for desktop environments and its use of raw screenshots, without the need for structured HTML or DOM data, significantly boosts its performance.

Among the specialized models, Fuyu (8 billion parameters) and CogAgent (18 billion parameters) show decent performance but remain uneven across the tasks. Fuyu scores well on Desktop-Text (33.0\%) but struggles in other areas, achieving an overall average of 15.3\%. CogAgent, with 74.2\% on Desktop-Text, stands out in specific tasks but falls short in overall performance with an average of 35.8\%. SeeClick, with 9.6 billion parameters, performs strongly with 72.2\% on Desktop-Text and an average of 39.3\%, though it struggles with WinSpot, where our model, WinClick, excels.

The key advantage of WinClick is its ability to outperform not only larger models like Fuyu and CogAgent, but also its own base model, Phi-3 Vision. Despite both models having the same number of parameters (4.2 billion), WinClick's specialized fine-tuning and data alignment approach provide a substantial performance boost. The results show that WinClick, particularly when fully fine-tuned, provides the best overall performance, demonstrating that task-specific fine-tuning on GUI-related tasks leads to significant improvements over a more general base model. Additionally, our model's ability to operate without relying on structured data like HTML or DOM files makes it highly effective for desktop GUI environments, where such data is often unavailable. These results confirm the superiority of our approach in developing GUI agents for Windows applications.

\section{Error Analysis}

In this section, we perform a detailed error analysis of the incorrect predictions made by WinClick on the \textit{WinSpot} benchmark. While our model achieved a correct prediction rate of 56.2\%, 461 out of 1,052 total predictions were incorrect. To better understand these errors, we categorized the incorrect predictions based on the distance between the predicted points and the ground truth (center of the target element).
Table~\ref{table:error_analysis} presents the distribution of incorrect predictions by distance. Most errors are found to be relatively close to the target, with over 60\% of the incorrect predictions within 0.2 units of the ground truth. This suggests that the model generally identifies the correct region but struggles with fine-grained localization. However, around 10\% of the cases have errors exceeding 0.6 units, indicating that in some instances, the model completely misinterprets the visual context or instruction.
These findings point to two main areas for future improvement: enhancing fine-grained localization to reduce errors close to the target and addressing the outlier cases where predictions are far from the intended target. Focusing on these areas could significantly improve the overall accuracy of the model.

\begin{table}[h]
\centering
\scalebox{0.9}{
\begin{tabular}{lcc}
\hline
\textbf{Distance (in units)} & \textbf{Incorrect} & \textbf{Total(\%)} \\ \hline
0.0 - 0.1               & 107                           & 23.2\%                      \\ 
0.1 - 0.2               & 177                           & 38.4\%                      \\ 
0.2 - 0.3               & 84                            & 18.2\%                      \\ 
0.3 - 0.4               & 52                            & 11.3\%                      \\ 
0.4 - 0.5               & 29                            & 6.3\%                       \\ 
0.5 - 0.6               & 9                             & 2.0\%                       \\ 
>0.6                    & 43                            & 9.3\%                       \\ \hline
\textbf{Total}          & 461                           & 100\%                       \\ \hline
\end{tabular}}

\caption{Distribution of Incorrect Predictions based on Distance from Ground Truth.}
\label{table:error_analysis}
\end{table}

\noindent In summary, while WinClick demonstrates strong localization capabilities, fine-tuning is required to improve precision for errors close to the target and to reduce large errors in outlier cases. Addressing these limitations will enhance the model's performance in complex GUI tasks.

\section{Conclusion}
In this paper, we presented WinClick, a novel visual GUI agent designed to automate desktop UI interactions using raw screenshots without relying on structured data formats such as DOM or HTML. Our approach successfully addresses the challenges of GUI grounding in desktop environments like Windows, and our model outperforms baseline methods on both the WinSpot and ScreenSpot benchmarks, demonstrating its effectiveness in complex, real-world GUI tasks.
We also introduced WinSpot, a new benchmark specifically tailored for evaluating GUI agents in desktop environments. This benchmark will serve as a valuable resource for further research in GUI automation. Going forward, we see significant potential in expanding WinClick to handle more dynamic UIs, incorporate user interaction data, and refine its fine-tuning processes to improve accuracy and efficiency. WinClick lays a strong foundation for future advancements in desktop GUI automation, paving the way for more scalable and adaptable solutions in this growing field.
\section{Future Work}

While WinClick demonstrates significant advancements in desktop GUI automation, several areas offer opportunities for further exploration. One is improving fine-grained localization in dense or visually complex UIs, where elements may be tightly packed or visually similar. To address this, future work could explore techniques such as iterative refinement, where the model makes successive adjustments to its predictions, or by incorporating contextual information about the overall UI layout to help the model prioritize certain regions.
Handling dynamic UIs is another important direction. Currently, WinClick operates on static screenshots, which limits its ability to interact with interfaces that change based on user input or system state. Leveraging techniques like reinforcement learning could be a promising approach, allowing the model to learn from its actions and refine its interaction strategies over time, making it more adept at handling dynamic, evolving UIs.
Additionally, expanding WinClick to support multi-step workflows, where the model needs to navigate through a sequence of actions rather than just responding to isolated tasks, could improve its applicability in complex software environments. This would involve the model maintaining an internal state or memory, which could help it better understand long-term tasks and dependencies between different GUI elements.
For the WinSpot dataset, we plan to collect a more diverse set covering a broader range. The ultimate goal is to develop an agent capable of functioning across all GUIs, across all operating systems.

\clearpage
\section{Limitation}

WinClick has several key dependencies for successful implementation. It relies on high-quality data for training, as constructing a large, diverse dataset of screenshots with detailed interaction labels is essential. Additionally, training the model requires significant computational resources, even with the smaller Phi3-vision model, which may pose challenges for organizations with limited access to such infrastructure. Finally, WinClick is designed specifically for Windows, limiting its generalizability to other platforms like macOS or Linux.
\bibliography{custom}
\appendix

\section{Training Details}
\label{train}
To improve the model's understanding of GUI images, we used both full fine-tuning and LoRA to explore efficient parameter updates, and also updated Visual encoder. For full fine-tuning,  the visual encoder with setting the initial learning rate to 2e-6. The initial learning rate for phi3 was 5e-6, batch size was 32 and a warmup ratio of 0.03. The training process use 4 * NVIDIA H100 GPUs.

\section{More Training Data Construct Examples}
\label{moreexample}
\begin{figure*}[htbp]
  \begin{center}
  \includegraphics[width=1.9\columnwidth]{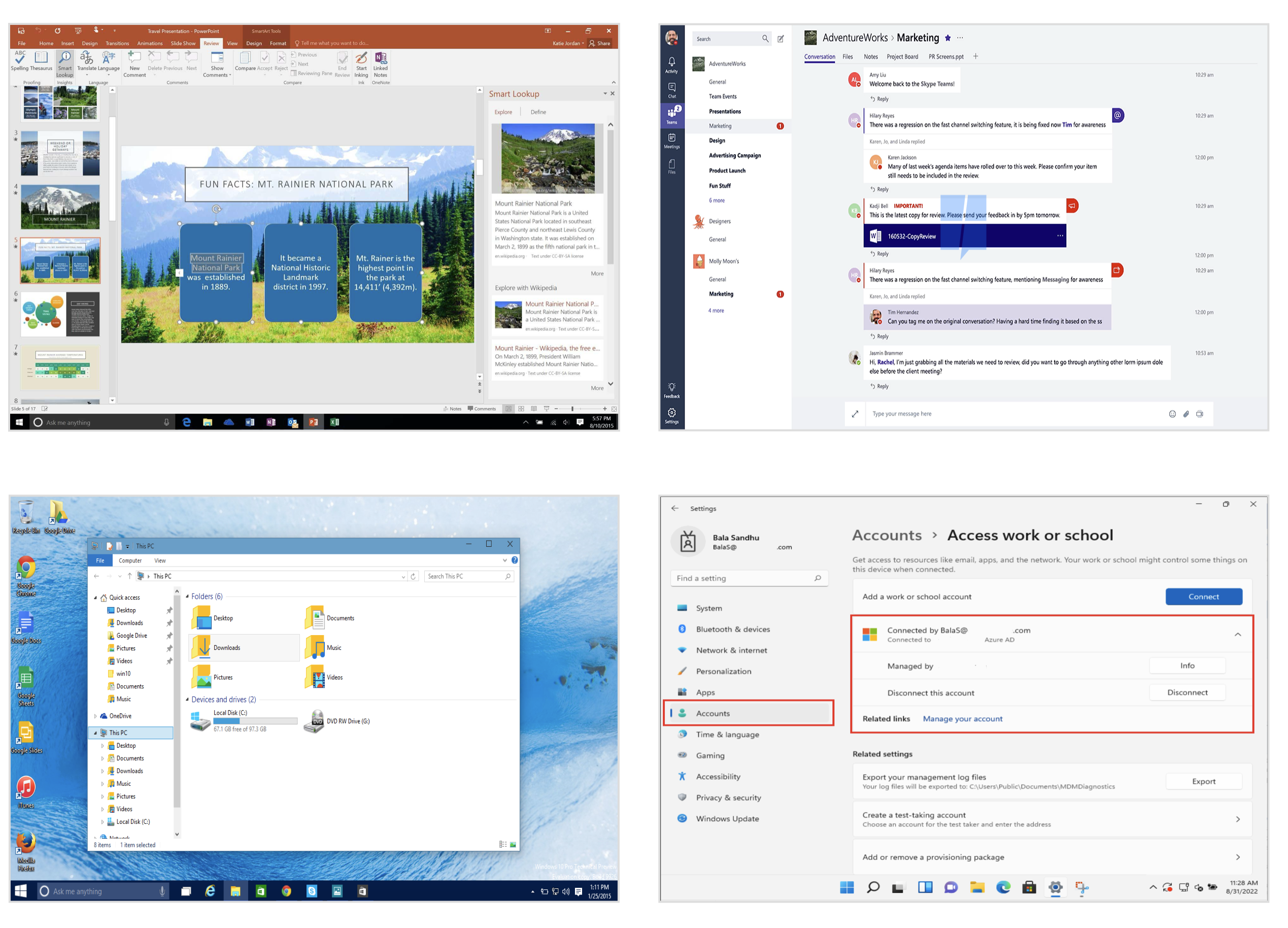}
  \end{center}
  \captionsetup{justification=centering}
  \caption{Training Data examples }

  \label{fig:ap1}
\end{figure*}

More data examples are shown in Figure \ref{fig:ap1}.

\section{Human Annotation}
\label{human}
The annotation process for WinSpot involved a group of carefully selected annotators, all of whom were undergraduate, master’s, or Ph.D. students, proficient in GUI operations and familiar with the Windows operating system. The annotation team consisted of individuals with diverse academic backgrounds, ensuring a broad understanding of GUI interactions across different applications. Each annotator was tasked with identifying and marking interactable regions within various Windows applications, focusing on elements such as buttons, icons, menus, and other clickable UI components.
For the annotation process, annotators were provided with a set of Windows screenshots. These screenshots were then annotated using a custom tool that allowed them to create bounding boxes around each interactable element. Annotators were also required to provide corresponding descriptions of the elements, ensuring that both the visual and functional aspects of each UI component were documented. The entire annotation process was conducted in English to maintain consistency across all samples.
To ensure data privacy, all screenshots were reviewed and post-processed to remove any personal information or sensitive content.  The final dataset includes over 1,000 images and 5,000 instruction-click pairs, representing a comprehensive set of interactions across a variety of Windows applications.
\end{document}